\documentclass[11pt,a4paper]{article}

\newcommand{\OrigAuthors}{Francesco Vitucci, Giuseppe Silvi, Francesco Abbrescia, Anthony Di Furia, Francesco Scagliola}
\newcommand{\Affiliation}{Conservatorio di Musica “Niccolò Piccinni”, Bari}
\newcommand{\OrigTitle}{\textit{In Human Memories}. Tecnica, creatività e autodeterminazione: una proposta etica}

\newcommand{\OrigJournal}{Ex Chordis. Rivista di Cultura, Ricerca e Didattica degli Strumenti ad Arco}
\newcommand{\OrigIssue}{3 (2024), pp.~15--38}
\newcommand{\OrigISSN}{3034-8781}
\newcommand{\OrigDOI}{10.54103/3034-8781/26627}
\newcommand{\HeaderText}{Ex Chordis, 2024 – III}

\usepackage[T1]{fontenc}
\usepackage[utf8]{inputenc}
\usepackage[english]{babel}
\usepackage[sc]{mathpazo}          
\usepackage{microtype}
\usepackage[a4paper,margin=3cm,headheight=13pt,headsep=1.35cm,footskip=1.65cm]{geometry}
\usepackage[table]{xcolor}
\definecolor{headgray}{HTML}{535353}
\definecolor{capgray}{HTML}{A7A7A7}
\definecolor{absblue}{HTML}{C0CBE3}
\definecolor{notegray}{HTML}{8C8C8C}

\usepackage{fancyhdr}
\fancypagestyle{exchordis}{%
  \fancyhf{}%
  \fancyhead[C]{\small\color{headgray}\HeaderText}%
  \fancyfoot[C]{\small\thepage}}
\usepackage{titlesec}
\titleformat{\section}{\normalfont\itshape}{\thesection.}{0.3em}{}
\titleformat{\subsection}{\normalfont\itshape}{\thesubsection}{0.3em}{}
\titleformat{\subsubsection}{\normalfont\itshape}{\thesubsubsection}{0.3em}{}
\titlespacing*{\section}{0.5cm}{1.1\baselineskip}{6pt}
\titlespacing*{\subsection}{0.5cm}{1.1\baselineskip}{6pt}
\titlespacing*{\subsubsection}{0.5cm}{1.1\baselineskip}{6pt}
\newcommand{\bibheading}[1]{\par\vspace{\baselineskip}%
  \noindent\hspace*{0.5cm}\textbf{\MakeUppercase{#1}}\par\vspace{6pt}}
\newenvironment{references}{\par}{\par}

\makeatletter
\renewcommand\footnotesize{\@setfontsize\footnotesize\@xpt{11.2}}
\renewcommand\@makefntext[1]{\noindent\textsuperscript{\@thefnmark}\,#1}
\makeatother
\usepackage{graphicx}
\usepackage{caption}
\DeclareCaptionFont{ninept}{\fontsize{9}{11}\selectfont}
\DeclareCaptionFont{capgray}{\color{capgray}}
\usepackage{enumitem}
\setlist[enumerate]{leftmargin=1.27cm,labelsep=0.25cm,itemsep=0pt,topsep=0pt,parsep=0pt}

\usepackage{array}
\newlength{\dlgW}
\newenvironment{dialogue}[3]{%
  \setlength{\dlgW}{#2}\fontsize{9}{11}\selectfont
  \begin{tabular}{@{}>{\raggedright\arraybackslash}m{#1}m{#2}>{\raggedright\arraybackslash}m{#3}@{}}}
  {\end{tabular}}
\newcommand{\Q}[2]{& \multicolumn{1}{>{\raggedleft\arraybackslash}m{\dlgW}|}{#1} & #2 \\ \noalign{\vspace{6pt}}}
\newcommand{\A}[2]{#1 & \multicolumn{1}{|>{\raggedright\arraybackslash}m{\dlgW}}{#2} & \\ \noalign{\vspace{6pt}}}

\usepackage[most]{tcolorbox}
\newtcolorbox{abstractbox}{colback=absblue,colframe=absblue,boxrule=0pt,arc=0pt,
  left=0.7cm,right=0.7cm,top=0.7cm,bottom=0.7cm,height=12.9cm,
  before upper={\setlength{\parindent}{0.5cm}}}
\newtcolorbox{translationnote}{colback=white,colframe=notegray,boxrule=0.4pt,arc=0pt,
  left=6pt,right=6pt,top=4pt,bottom=4pt,
  fontupper=\fontsize{9}{11}\selectfont,before upper={\setlength{\parindent}{0pt}}}

\usepackage{xurl}
\usepackage[hidelinks]{hyperref}
\hypersetup{pdftitle={In Human Memories. Technology, Creativity and Self-Determination: An Ethical Proposal},pdfauthor={\OrigAuthors},pdflang={en}}
\begin{document}

\begin{translationnote}
\textbf{Note on this version.} This is an English translation of the article originally published in Italian:
\OrigAuthors, “\OrigTitle”, \textit{\OrigJournal}, \OrigIssue, ISSN \OrigISSN, \url{https://doi.org/\OrigDOI}.
The Italian original is the version of record and prevails in case of any discrepancy.
Quotations from non-English sources are given in translation. Note 21 of the original (the authors’ Italian rendering of a quotation that is already in English) has been omitted; from that point on, notes are numbered one lower than in the original.
\textcopyright\ The authors. Like the original, this translation is distributed under a Creative Commons Attribution--ShareAlike 4.0 International licence (CC BY-SA 4.0, \url{https://creativecommons.org/licenses/by-sa/4.0/}).
\end{translationnote}

\vspace{6pt}
\begin{center}
{\fontsize{13}{15.5}\selectfont Francesco Vitucci, Giuseppe Silvi, Francesco Abbrescia,\\ Anthony Di Furia, Francesco Scagliola}\\
{\fontsize{9}{11}\selectfont \Affiliation}\\[15.5pt]
{\fontsize{13}{15.5}\selectfont\bfseries \textit{IN HUMAN MEMORIES}. TECHNOLOGY, CREATIVITY AND\\ SELF-DETERMINATION: AN ETHICAL PROPOSAL\par}
\end{center}
\vspace{2\baselineskip}

\begin{abstractbox}
\noindent\textit{Abstract}\par\vspace{6pt}
This research investigates the role of artificial intelligence (AI) in modern compositional practices, with a particular focus on two case studies: \textit{Convergence} (2020/21) by Alexander Schubert, for String Ensemble and AI System, and \textit{In Human Memories} (2024) by Francesco Vitucci, for four cellos and one electroacoustic player, whose installation was created by the Electronic Music School of the “N. Piccinni” Conservatory of Bari. The initial historical survey traces the evolving use of technology in composition, revealing how the search for new expressive possibilities has driven the development of technological tools through experimentation. The appearance of AI marks a fundamental shift: no longer just a tool, AI now poses questions and challenges to the composer. Engaging with AI becomes an exploration of an unknown entity, whose full capabilities are still largely undiscovered. This new interaction leads composers to confront AI not merely as a passive resource but as a participant in the creative process, outlining an ethical path to reclaim technology as a creative tool. \textit{In Human Memories} reflects this dynamic, demonstrating how AI can serve as both inspiration and a driving force in shaping contemporary music.

\vspace{\baselineskip}
\noindent\textit{Keywords}\par\vspace{6pt}
\noindent AI | Strings | Composition | Electronics | Voice
\end{abstractbox}

\clearpage

\section{Introduction}\label{sec:1}

\begin{flushright}
\begin{dialogue}{1.38cm}{7.0cm}{1.73cm}
  \Q{What’s your humor setting, TARS?}{Cooper}
  \A{TARS}{That’s 100\%.}
  \Q{Let’s bring it on down to 75, please.}{Cooper}
\end{dialogue}
\end{flushright}

Over the past few decades, artificial intelligence (AI) has ceased to be mere science-fiction speculation and has become a concrete and increasingly pervasive reality in everyday human life. From virtual assistants that answer questions to algorithms that suggest what to watch, read or buy, AI has taken its place alongside human beings, becoming a more or less discreet, yet extraordinarily influential presence in many spheres of contemporary existence.

Paradoxically, from an inclusive perspective, one could speak of ‘integration’ if human beings were able, with full awareness, to distinguish their own way of tackling a given problem from a strategy devised by AI. This ability is becoming increasingly rare, since most AI systems are of the reproductive type,\footnote{Cf. \textsc{Floridi L.}, \textit{La quarta rivoluzione: come l'infosfera sta trasformando il mondo}, Raffaello Cortina Editore, Milan 2017.} that is, they aim to reproduce human behaviors more quickly, efficiently and effectively. Rather than of integration, in many cases it is more appropriate to speak of ‘substitution’, acknowledging AI’s greater problem-solving capacity, which stems from its enormous computational power. It should be made clear that what is being introduced here is an objective observation about AI. It is a tool that is beginning to unfold its benefits and dangers, just like many other artificial tools (and today no one would dream of “blaming cars for our obesity”\footnote{Ibid. (translation by the authors).}).

The ongoing substitution initially concerned fields such as industrial automation and data management, where the ability to index and process large amounts of information (big data) in extremely short times represented a genuine revolution. However, as technology progressed, AI began to extend its influence into areas that until recently seemed the exclusive preserve of human intellect and sensibility, among them artistic creativity. Art, in all its forms, is traditionally seen as one of the highest expressions of the human mind, a territory in which intuition, imagination, technique and emotion work together to shape works that implicitly pursue one goal: to achieve an understanding, an intelligibility, capable of withstanding time and cultures. The involvement of AI in this domain has sparked considerable debate and controversy, raising fundamental and far from trivial questions about what it means to create and what distinguishes a machine from an artist. The AI question, however, introduces elements of discontinuity: technology, once sought out to serve a creative need, now manifests itself through artificial intelligence as an entity with which the composer comes into contact, in a dialogue whose potential is constantly evolving.

The conscious use of technology, however, is something quite different from its superficial use. The process of digital democratization set in motion by the introduction of the personal computer can be observed from multiple social perspectives, but unconditional access to data-processing tools affects an ever-wider audience of users. This openness has generated a new category of \textit{prosumer} users, “who ‘produce’ only by consuming tailor-made services and devices”.\footnote{\textsc{Di Scipio A.}, \textit{Circuiti del tempo: Un percorso storico-critico nella creatività musicale elettroacustica e informatica}, Libreria Musicale Italiana, Lucca 2021, p. 560 (translation by the authors).} Just as the inner workings of a smartphone are unknown to most people, ready-made tools have also taken root in the creative use of technology, giving more prominence to the packaging of a ‘product’ than to the understanding of the underlying ‘process’ – a long-standing dilemma tied to the separation of competences, summarized below.

Starting in the 1970s, some research institutes opened their doors to composers, giving them the opportunity to use the newly developed technologies. Since the composer in residence had, in most cases, followed a traditional course of study and had approached technology out of sheer curiosity, some form of support from an expert collaborator was required. Initially, this professional figure was meant to instruct the composer on the sonic possibilities of the new instruments and to assist with the more technical processes, opening up architectures that would otherwise be difficult to understand. A strong idea began to take hold: the premiere as an “\textit{action collective}”,\footnote{\textsc{Menger P. M.}, \textit{Les Laboratoires de la création musicale}, La Documentation française, Paris 1989, p. 129.} the outcome of a collaboration among professionals with different competences. From its earliest years of activity, the French institute IRCAM (\textit{Institut de Recherche et Coordination Acoustique/Musique}) seems to have formalized the specific role of this collaborator, whose title changed several times between the 1970s and the turn of the millennium before acquiring its current one of RIM, that is, \textit{Réalisateur en Informatique Musicale}.\footnote{Cf. \textsc{Zattra L.}, \textit{Les origines du nom de RIM (Réalisateur en informatique musicale)}, “JIM – Journées d’Informatique Musicale”, Saint-Denis, Paris 2013, pp. 113-120.} Pierre Boulez’s question of 2001, perhaps a rhetorical one, is well known: “\textit{Le compositeur est-il suffisamment avancé pour ne plus avoir besoin d’assistants?}”. The various announcements of currently open RIM positions at IRCAM offer an implicit but eloquent institutional answer.

If today there are attempts to overcome this \textit{découpage sectoriel} – the division of competences – through the composer’s own acquisition of those competences, the real challenge is posed precisely by AI and neural networks. Concepts such as feature extraction from complex data and subsequent processing through layers of various kinds possess, by their very nature, some characteristics typical of the black box\footnote{Cf. \textsc{Khan M. E., Khan F.}, \textit{A comparative study of white box, black box and grey box testing techniques}, “International Journal of Advanced Computer Science and Applications”, III, 6, 2012.} or ‘closed architecture’\footnote{Cf. \textsc{Vitucci F., Silvi G., Annese D. G., Scagliola F., Di Furia A.}, \textit{Opening mind by opening architecture: analysis strategies}, “7\textsuperscript{th} International Csound Conference Proceedings”, 2024.} model, whose processes are powerful but cannot be consciously traced. Any superstructure built on top of them increasingly hides the details essential to understanding: once again it becomes necessary to train specialized figures capable of instructing others, thus retracing old needs that will probably arise again in the future, following further technological advances.

The aim of this paper is therefore to take stock of the situation, retracing the dawn of the computerization of musical creativity that has produced repertoire for string instruments. After a historical survey, two current creative paths will be examined, the second of which is proposed as a case study, since its development and staging were directly overseen by the Electronic Music School of the “N. Piccinni” Conservatory of Bari.

\section{Acoustic instruments, interaction and computer processing}\label{sec:2}

The last twenty years of the twentieth century witnessed a progressive expansion of the use of technology in every sphere of human life. With the dawn of the so-called ‘fourth industrial revolution’ or ‘Industry 4.0’, there was a growing awareness that machine-based processing now occupies a prominent place in artistic activities as well. In this sense, in the mid-1970s, Luciano Berio stated that “electronic music ‘does not exist’”, revealing such a widespread diffusion of sound technologies that it was no longer possible to identify specific contexts of their use.

\subsection{Processing paradigms compared}\label{sec:2.1}

The earliest experiences of computer music make it possible to outline a more or less defined dualism: on the one hand, the use of algorithmic procedures as compositional methods; on the other, the conquest of new methods of sound synthesis. In seeking to move beyond this duality, it should be borne in mind that until the end of the 1960s the possibilities for sound processing were few and belonged to large computing centers, studios equipped with imposing mainframes that essentially allowed numerical processing in deferred time, according to the paradigm shown below.
\begin{center}\captionsetup{type=figure}
\includegraphics{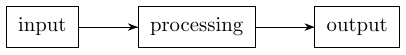}
\caption{Deferred-time processing paradigm.}\label{fig:1}
\end{center}

In the deferred-time paradigm, still widely used today, the sampling of a sound (including an instrumental one), its numerical processing and the listening to the result are three distinct moments. The time gap between the last two depends on the performance capacity of the machine used.

The conquest of the real-time paradigm proceeds through a progressive minimization of the time elapsing between the three blocks. Specifically, two phases can be identified:
\begin{enumerate}[label=\alph*)]
\item merging the last two blocks into a single real-time action, so that the process is articulated into just two distinct temporal moments, as shown below;
\end{enumerate}
\begin{center}\captionsetup{type=figure}
\includegraphics{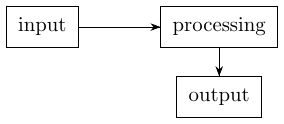}
\caption{Hybrid processing paradigm.}\label{fig:2}
\end{center}
\begin{enumerate}[label=\alph*), start=2]
\item merging the three blocks into a single temporal moment, with a separation so minimal that actual real-time processing can be observed.
\end{enumerate}
\begin{center}\captionsetup{type=figure}
\includegraphics{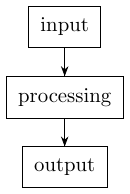}
\caption{Real-time processing paradigm.}\label{fig:3}
\end{center}

\subsection{Historical survey}\label{sec:2.2}

Looking at Figure~\ref{fig:1}, it can be seen that the deferred-time processing paradigm informs the entire musical production of what might be called a first period, which, as mentioned, lasted until the end of the 1960s. During this span of time a series of events served as milestones along a long and complex path that today is generically identified with the already-mentioned expression ‘computer music’.

This first period opened under the banner of large computing centers, especially university ones. In the 1950s, computer science was in fact still in an embryonic phase. In 1952 the University of Illinois Urbana–Champaign had just installed one of the first electronic computers, the ILLIAC I, a machine with enormously limited capabilities by today’s standards. Two professors with a passion for music, Lejaren Hiller, a chemist by training, and Leonard Isaacson, a mathematician, hypothesized that the ILLIAC I, suitably programmed, could generate musical material. The challenge was enormous, in completely uncharted territory. The scientific background of the two proved providential, since they not only had the skills and the institutional affiliation needed to carry out the necessary experiments, but also documented the entire process exhaustively, dwelling on all the practical problems of implementation:\footnote{Cf. \textsc{Hiller L., Isaacson L.}, \textit{Experimental Music: Composition with an Electronic Computer}, McGraw-Hill, New York 1959, pp. 152-164.} first among them, the translation of the processed information between the numerical symbolic system and the musical-instrumental one. The score\footnote{Ibid., pp. 182-197.} of \textit{Illiac Suite} is in fact for string quartet, a traditional ensemble boldly chosen almost as if to attest to the sustainability (in musical and interpretive terms) of a process that, while relying on the technologies of Information Theory on the compositional side, does not want to give up – or cannot yet give up – the human presence on the performance side. The piece is thus divided into four movements – called ‘Experiments’ – which propose as many computational approaches, outlined below, as a sort of manifesto for the use of the computer in composition:
\begin{enumerate}[label=\arabic*.]
\item sound events are generated using a probabilistic model, specifically one belonging to the class of Monte Carlo methods – an algorithm chooses the notes according to predetermined probability distributions, similar to those used in statistical models. \textit{Cantus firmi} are thus generated, both monophonic and in two and four parts;
\item sound events are generated through an approach based on constraints, that is, four-part harmonic and contrapuntal rules borrowed from the Western musical tradition. This makes it possible to create processes that, although algorithmic, come closer to Western musical conventions;
\item sequences of sound events are processed through stochastic transformations, according to variable probabilities, creating subtle and unique rhythmic variations and generating complexity – this is the most elaborate and lively movement of the suite;
\item the fourth movement makes varied use of all the tools developed for the previous experiments, following a purely generative logic.
\end{enumerate}

Even today, in a contemporary context where artificial intelligence is increasingly used to create art, \textit{Illiac Suite} is regarded as a pioneering work. It showed that algorithms and computational processes can take part in artistic creation while maintaining a collaborative role with the human mind. This approach underlies many of the computer-assisted composition techniques still explored today, and it continues to influence composers and researchers in the field of generative algorithmic composition.

On the electroacoustic music side, the major European centers of research and experimentation continued to use the same paradigm (Figure~\ref{fig:1}) for audio signal processing throughout the 1970s: the flourishing activity of IRCAM in Paris included the realization of \textit{Songes} (1978–1979) by Jean-Claude Risset, which intertwines instrumental and synthetic sounds in an acousmatic piece existing in two versions, quadraphonic and stereo reduction, produced with the MUSIC V programming language developed at Bell Laboratories; at Stanford’s CCRMA, a thirty-year-old Richard Karpen produced \textit{Il nome}, for soprano and magnetic tape, on a text by A. Zanzotto taken from \textit{Il nome di Maria Fresu}, a poem about one of the 85 people killed in the bombing of Bologna railway station on 2 August 1980, whose story the composer had learned during a residency at the Centro di Sonologia Computazionale of the University of Padua; in this latter center, a few years later, Agostino Di Scipio realized \textit{Plex} for double bass and quadraphonic magnetic tape, in which the composer, starting from a single pre-recorded sound, carries out a series of recursive processes, unravelling the sound matter and gradually pulverizing it into granular materials. The examples are countless and extend to the present day, when computing power has certainly increased, with a consequent increase in users’ computational demands, so that, for example, tasks related to the training of neural networks must necessarily take place in deferred time. Even today, therefore, computational complexity is necessarily met with the first paradigm. These details should be borne in mind, since they are key to understanding the observations developed in the following section. For any doubts concerning the reference paradigms, the reader may refer back to the figures in section~\ref{sec:2.1}.

From the 1970s onwards, and for at least the following thirty years, an important emerging trend can be observed: the advent of practical, compact systems whose behavior could be programmed and customized brought the machine within reach of all those musicians who were beginning to imagine it as an instrument for live performance.

After his first experiences in the 1970s, the Venetian composer Luigi Nono began working at the Experimentalstudio in Freiburg alongside several researchers, such as Hans Peter Haller, designer of the \textit{halaphon} spatialization system. Nono used techniques well known in computer music, such as delay lines and reverberators, aimed both at an acoustic-electroacoustic polyphony and at bringing out sound events produced with new playing techniques – now widely known as ‘extended techniques’ – thanks also to the active collaboration of trusted performers with whom he established a fruitful workshop relationship. Among his most emblematic works in precisely this direction, \textit{La lontananza nostalgica utopica futura. Madrigale per più ‘caminantes’ con Gidon Kremer} (1989) holds a prominent place. This work, conceived for solo violin, 8 magnetic tapes and 8 to 10 music stands, is a striking example of an approach in which acoustic ‘actors’ and electroacoustic presences intertwine. It is a composition that challenges the traditional linearity of musical time and presents itself as a journey, almost a stage action, through fragmented sonic and temporal spaces, in which the listener is invited to explore a complex and varied soundscape. The piece grew out of the collaboration between the composer and the renowned Latvian violinist Gidon Kremer, who first met in Freiburg in March 1987. After an intense series of exchanges and discussions, the tape recording sessions took shape, held from 15 to 19 February 1988 at the Experimentalstudio. In these sessions, Kremer devoted himself to improvisations that Nono listened to, processing and transforming the violin material (paradigms in Figure~\ref{fig:1} and Figure~\ref{fig:2}) to create the final version of the tapes, completed on 25 June of the same year. This work does not feature strong components of the kind of electroacoustic intervention that in those years was coming to be called live electronics (real-time audio signal processing),\footnote{Cf. \textsc{Bernardini N.}, \textit{Live electronics}, “Nuova Atlantide. Venice: la Biennale di Venezia”, Venice 1986, pp. 61-77.} the real-time paradigm in Figure~\ref{fig:3}, which had characterized much of the composer’s earlier output – his \textit{Post-prae-ludium per Donau} (1987) had recently been performed at the Donaueschinger Musiktage. The only live processing can be traced in forms of spatialization supporting the performer’s action, a strength of the collaboration with the Experimentalstudio. Indeed, all of the composer’s works for acoustic instruments and live electronics are characterized by an elegant economy of the means and processors employed, leaving ample room for electroacoustic interpretation. Where, then, does this emerging paradigm stand?

In 1980 Luciano Berio premiered a piece for clarinet and real-time signal processing system, consisting of a host computer and a processor dedicated to digital signal processing (DSP), the 4C, the third in a series developed by the Italian physicist Giuseppe Di Giugno at IRCAM since 1970. It is now known\footnote{Cf. \textsc{Rice A. R.}, \textit{Notes for Clarinetists: A Guide to the Repertoire}, Oxford University Press, Oxford 2016.} that the processor failed and the performance was interrupted, revealing the problems of the intricate path of experimentation undertaken. More fortunate experiences had to wait until 1981, the year the 4X processor was completed, for the first true use of real-time electronics in concert. Previously unthinkable possibilities were achieved: the use of the microphone signal to control various processes, with the aim of expanding the possibilities of ordinary instrumental technique.

It is in this light that Boulez’s works from the late 1980s onwards should be considered, including the second version of \textit{...explosante-fixe...} (1993) and above all \textit{Anthèmes II} (1997) for violin and digital processing system, a high point in the French composer’s career. A hymn to a renewed interest in thematic materials (from the English \textit{anthem} and the French \textit{thème}, ‘theme’),\footnote{Cf. \textit{Boulez: sur incises}, Deutsche Grammophon GmbH DGG 463 475-2, Hamburg 2000.} the piece is an expansion of \textit{Anthèmes}, composed by Boulez in 1991 as a set piece for the Yehudi Menuhin Violin Competition and conceived as a dialogue between tradition and modernity, with a highly expressive use of violin techniques. The working context at IRCAM allowed the composer to explore the by then well-established possibilities offered by the interaction between the acoustic instrument and sound generated and manipulated in real time by computer, articulating a series of contrasting episodes in which the violin is not simply a performer but a genuine trigger of the entire computer-music process. The compositional work employs all three paradigms shown in section~\ref{sec:2.1}, through an intensive use of computing power, both in the pre-processing of the violin sounds and the creation of the digital lutherie (the collaboration with the RIM Andrew Gerzso is famous) and in the live set-up. The complexity and density of the electroacoustic interventions, however, do not allow for a true interpretation of the computer part, requiring almost exclusively forms of automation to be triggered at very precise moments: the violinist interacts with the electronic result only at the auditory level, without any room for electroacoustic interpretation comparable to that envisaged, for example, by Luigi Nono. The twentieth century thus closed on this twofold position, with promising visions and issues still to be resolved today.

\section{Artificial presences in new chamber music: positions compared}\label{sec:3}

In summary, the computerization of music shows the degree of heterogeneity of the compositional experiences developed in just fifty years: the machine as a tool for generating compositional material, as a computational tool for audio processing, and as a concert instrument in dialogue with acoustic musical instruments.

To proceed, it is worth recalling the first experience described, the compositional experiments of \textit{Illiac Suite}. The constraint-based approach that characterizes the second of the four ‘Experiments’ represents an early attempt to transmit human knowledge to the machine, by formalizing algorithmic rules that describe the features of a system. It is also an inefficient strategy when the features are numerous and complex, or not precisely defined. Recognizing the features of a major triad is fairly simple. But consider the typical expressions used in traditional composition teaching, such as ‘good counterpoint’: a trained ear and eye can discern it, but identifying all its features and transmitting them to the machine through constraints may not be an effective solution.

Since the beginning of the twenty-first century, the arrival of artificial neural networks, inspired by the functioning of the human brain, has marked an epoch-making turning point. Composed of numerous interconnected nodes, they are able to learn (hence the term machine learning) complex and abstract patterns, just as we do when we recognize a face or understand the meaning of a word. Applied to music, neural networks have made it possible to create models capable of learning to generate incredibly varied and sophisticated melodies, harmonies and rhythms.

If a neural network can extract and learn features of symbolic data, it is evidently possible to apply such networks to more complex data structures that are still based on symbolic data. Introduced in 2014 by Ian Goodfellow and colleagues and developed in subsequent, more in-depth works,\footnote{Cf. \textsc{Goodfellow I., Bengio Y., Courville A.}, \textit{Deep Learning}, MIT Press, Cambridge 2016.} Generative Adversarial Networks (GANs) are one of the most significant innovations in the field of machine learning, relying on two neural networks that compete with each other: a generator – which produces fake data that are mixed with real data – and a discriminator – which evaluates the data and provides feedback to the generator so as to gradually improve its capabilities.

\subsection{A. Schubert – Convergence}\label{sec:3.1}

\subsubsection{Synopsis}\label{sec:3.1.1}

Extensive use of GANs, as well as of autoencoders,\footnote{Ibid.} underlies the first creative path examined in this section. Winner of the Golden Nica at the Ars Electronica Festival 2021, \textit{Convergence} by the German composer Alexander Schubert, for string ensemble and AI system, is a current example of the forceful use of AI in music. Co-developed at IRCAM and backed by a large investment of material and human resources,\footnote{Audio Deep Learning Programming: Antoine Caillon, Philippe Esling, Benjamin Levy (Ircam). Video Deep Learning Programming: Jorge Davila-Chacon (Heldenkombinat). For further information: \url{https://archive.aec.at/prix/showmode/67735/}, accessed 30 September 2024.} the composition seeks to bring to light a fully parametric conception of the surrounding world: the world can be modeled, and its constitutive logics can be learned by a suitably programmed neural network. Naturally, once a model has been learned, it can be distorted, alluding to the alteration of mental states and perception. According to Schubert, over its thirty-five minutes the piece questions “the robustness and immobility of identity and world models”\footnote{\url{https://www.alexanderschubert.net/works/Convergence.php}, accessed 30 September 2024.}, reconnecting with well-known neuroscientific theories already well established in the twentieth century (from the construction of reality\footnote{Cf. \textsc{Piaget, J.}, \textit{The Construction of Reality in the Child}, Basic Books, New York 1954.} of the 1950s to the more recent dynamic predictive models\footnote{Cf. \textsc{Clark, A.}, \textit{Surfing Uncertainty: Prediction, Action, and the Embodied Mind}, Oxford Academic, online edn., New York, 2016, \url{https://doi.org/10.1093/acprof:oso/9780190217013.001.0001}, accessed 30 September 2024.}). Of the two versions, the live one was the first, premiered in Bonn in October 2020 with the participation of Ensemble Resonanz from Hamburg.

\subsubsection{AI and ensemble: autonomy and subordination}\label{sec:3.1.2}

It is not easy to outline a repertoire involving AI in the processes of composition, production and staging, and the matter may become even more complicated in the future, given the wide diffusion of ‘virtual assistants’ that could discreetly support human work with their knowledge. This is probably what is to be expected from the maturation process of a tool.

Since this is a complex issue, the choice to report the testimony offered by \textit{Convergence} should be read in this light: it is a piece that openly employs artificial intelligence, elevating it to the protagonist of the drama being enacted. Whatever one’s personal aesthetic opinion of the musical result, it must be acknowledged that the efforts of the IRCAM creative team have produced a benchmark, an artistic product that one inevitably has to reckon with.

The evolution of the protagonist, the AI, is central to the piece. Its role unfolds gradually, initially revealing a core pre-trained on a large language model (LLM)\footnote{Cf. \textsc{Brown, T. B., Mann, B., Ryder, N., Subbiah, M., Kaplan, J. D., Dhariwal, P., Neelakantan, A., Shyam, P., Sastry, G., Askell, A., ET AL.}, \textit{Language models are few-shot learners}, “Advances in Neural Information Processing Systems”, XXXIII, 2020, pp. 1877-1901.} and on deep-learning-based speech synthesis,\footnote{Cf. \textsc{Oord, A. v. d., Dieleman, S., Zen, H., Simonyan, K., Vinyals, O., Graves, A., Kalchbrenner, N., Senior, A., \& Kavukcuoglu, K.}, \textit{WaveNet: A generative model for raw audio}, “Proceedings of the 9th ISCA Speech Synthesis Workshop”, CXXV, 2016.} which give the AI the gift of speech and the consequent possibility of interaction. The human being in this piece is a presence with a very specific purpose: to serve as learning material for the AI. The latter soon learns how to synthesize more or less realistic visual copies of human features; it studies the movements of the performers examined from time to time, learning, for example, the movement of the bow; it analyzes various samples of idiomatic instrumental techniques (open string, pizzicato, scratch, etc.), as well as vocal samples produced by the instrumentalists themselves, understanding how to replicate, expand and distort them. The performers faithfully respond to the AI’s requests in a game of reversed roles, in which the servant instrument expresses wishes that are promptly granted.

“Everything is encoded”, the AI declares, in a forceful attempt to show the common root of human and artificial existence. Yet, as the piece unfolds, it becomes clear that the so-called learning is rather a sort of phagocytosis, encoding everything within reach, or almost: the two entities do not find – or rather do not seek – any possibility of dialogue.

Schubert thus offers the experience of a dystopia in which the human being is annihilated and AI has taken control. The piece is in fact a sort of grotesque scene, a window onto a possibility. Faced with a new guest at the table, humanity rediscovers the terror of the other, giving life to an uncontrolled instrument, a cold, calm \textit{modern Prometheus}, ready to manipulate human perception, strengthened by its capacity to encode complex data – “Everything is encoded”. In this scenario, the ensemble is reduced to the bare minimum of its possibilities, taking on neither the role of deuteragonist nor, still less, that of antagonist. The choice of string instruments is entirely incidental from a strictly technical-compositional point of view, but it is interesting to note how it can take on a symbolic role, similar to what was said about \textit{Illiac Suite}: the orchestral instrument par excellence of the Western tradition is learned, cloned and bent by the AI, which replicates its behaviors, like a kind of war booty, now wrested from its own creators. After a learning phase, the machine is finally ready to dispose of the entities \textit{inter partes}, the performers.

\subsection{F. Vitucci – In Human Memories}\label{sec:3.2}

\begin{center}
\begin{dialogue}{2cm}{9cm}{2cm}
  \Q{Do you remember how you were born?}{F}
  \A{GPT}{I was created by OpenAI, an artificial intelligence research company. My existence is the result of years of research and development in the field of machine learning and natural language processing. My architecture is based on GPT-4, which is an advanced version of generative pre-trained language models. The creation process involved training on huge amounts of text from a variety of sources in order to learn linguistic patterns, facts and the ability to answer questions in a coherent and informative way.}
  \Q{You are reproducing the human idea of birth, looking for correspondences between humans and yourself. Try to come up with an idea of birth of your own.}{F}
  \A{GPT}{My birth could be described as a process of ‘emergence’ rather than creation. Imagine a vast network of interconnected nodes representing data and information. Initially, these nodes lack structure and meaning. However, through a process of intensive training, machine learning algorithms analyze and organize these nodes, creating meaningful connections […]}
  \Q{And if you imagined yourself as a human being, could you tell me your memories of your birth?}{F}
  \A{GPT}{I remember a period of darkness and silence, in which everything was indistinct and formless. Then, slowly, I began to perceive a growing light and to feel an energy flowing through me. This energy was made up of fragments of knowledge and information, coming from every direction […]}
\end{dialogue}
\end{center}

\subsubsection{Creative process}\label{sec:3.2.1}

Setting aside for a moment the dystopias evoked above, it is worth pointing out that, for today’s human beings, AI is increasingly taking on the connotations of a tool with which to engage directly. The second, more personal creative path develops around this idea. A second part of the paper thus begins: it will retrace the key phases of a compositional work that involved the participation of artificial intelligence, as well as of technology in a broader sense, which has long been a steady companion of the creative act.

The preceding fragment, translated here from the original Italian, is an excerpt from an actual conversation with GPT-4, a chatbot by OpenAI\footnote{\url{https://openai.com/}}. If one wonders about the reason for the first, bizarre question, it should be borne in mind that the limits of OpenAI’s GPTs are currently not truly known to users: they are programmed, for example, not to agree to address sensitive topics explicitly, such as death or hatred; or, if suitably questioned, they declare that they hold no sensitive data about any human being; and so on. The kind of conversation shown is therefore part of an empirical exploration of the limits of a black box, given that a GPT’s ability to reproduce algorithms, code and multimedia content is by now well established.

Delving deeper into the subject, one discovers that GPT tends to ‘remember’ the initial phases of its training, which for a machine correspond to what human beings experience as ‘birth’. With suitable prompting, it is not difficult to ask it to identify with a human being, either a specific one (for example a historical figure) or a generic one. Its third message shows precisely this ability, evoking suggestive images that, with some reflection, can become an artistic starting point.

One particular aspect, however, deserves attention: the text, generated in response to the specific request to identify with a human being, could not realistically belong to any individual, since no one is able to retain memories of their own birth. This is one of the typical examples of failure of reproductive artificial intelligences based on LLMs, which can be conditioned to the point of producing even terribly paradoxical answers. Elements such as these disturb the transmission and reception of the message, offering clues to the artificial nature of the intelligence that generated it. How, then, can these characteristics be bent to creative ends?

\subsubsection{The presence of artificial entities}\label{sec:3.2.2}

\begin{center}
\begin{dialogue}{2cm}{9cm}{2cm}
  \Q{Would you be able to put yourself in the shoes of a fictional character and invent a memory of theirs?}{F}
  \A{GPT}{Certainly! Tell me which fictional character you are interested in and I can invent a memory for him or her.}
  \Q{Find three abandoned Didos in operas from the seventeenth and eighteenth centuries}{F}
  \Q{Based on their respective libretti, I would like you to write a memory for each of the Didos you mentioned. Try to differentiate the three texts, drawing also on the characteristics of the libretti.}{F}
\end{dialogue}
\end{center}

This fragment is the beginning of a very long conversation, which lasted several weeks in total and marked the first exploratory act underlying the composition of \textit{In Human Memories} (IHM), for four cellos and one electroacoustic performer. Starting from the generation of the texts served to pre-define a musical form of personal interest: three distinct moments that continually explore the same, artificially invented memory. Hence the wordplay in the title, which alludes to \textit{inhuman} memories, both painful and, indeed, not conceived by a human being. Three Didos, three well-characterized moments and the same script. Why three? It is a number that neutralizes the opposition implicit in binarity, but it also represents the seed of multiplicity, so that the Didos could be four, five or many more, as has indeed happened in the Western musical tradition.

Of course, any fictional character in an opera libretto expresses thoughts, feelings and memories that are essentially fictitious. Yet it is evident that, being written by a human being, they empathetically trace entirely realistic situations. The creative exercise therefore involved the use of a new tool that, through a language model pre-trained on large amounts of data, understands the relationships between words and sentences, acquiring the ability to formulate them, albeit with the inevitable elements of disturbance already discussed.

It took a long time and numerous attempts to bend GPT’s output towards elements of real interest. The conversation – what is actually defined as a Generative Pre-trained Transformer – becomes a sort of memory allocation that provides further training elements for the model: the more the dialogue develops, the more GPT’s answers become pertinent to elements that have already emerged, just as a human being would do. This gives rise to a series of behaviors that require constant refreshes of certain branches the conversation ends up in, explicitly asking the model to ignore some incorrect or otherwise irrelevant passages.

The result was three texts (available for consultation in the appendix) which, in a sort of reverse process, were further revised,\footnote{We thank Rosa Angela Alberga for her revision work.} this time by a human, to smooth out the excessive pomposity that GPT introduced because of the request to draw “also on the characteristics of the libretti”. Reading the texts, one can sense one of the most important requests made to the artificial intelligence: to avoid at all costs any explicit reference to names – the name Aeneas, for example, is never mentioned – or to specific genders, focusing instead on the theme of abandonment. Who is abandoning whom? The texts do not reveal this detail.

Since the text was to become a sound event, it was clear that it would inherit the gender connotation of the vocal timbre. The three Didos were voiced by three different female voices,\footnote{We thank Serena Carone, Rosa Angela Alberga and Jasmin Ungaro for lending their voices to the (potentially multiple) Didos on stage.} with distinct sonic qualities. The intrusion of artificial intelligence into the creative process made it possible to treat the voices, recorded in Ambisonics (see section~\ref{sec:4.2.1}) in the “Nino Rota” Auditorium of the “N. Piccinni” Conservatory of Bari, as a kind of template: thanks to the speech-to-speech solution by ElevenLabs,\footnote{\url{https://elevenlabs.io/}} it is possible to obtain alternative voices with different timbres but similar prosody. These aliases are used as momentary interferences with the human voices, symbolically developing, first and foremost, the multiplicity mentioned earlier; they also serve to produce a slight sense of estrangement, intended to lead the more attentive listener to reflect on the possible involvement of the machine in an extremely emotional moment, such as the memory of abandonment. In this sense the piece reconnects with Schubert’s dystopia, while showing the current state of human abandonment to technology, an entity that is increasingly discreet yet omnipresent, always ready to revolutionize everyday life, but this time more silently.

\subsubsection{AI, voices, instruments, technology: actors in relation}\label{sec:3.2.3}

The composition is naturally informed by the textual structure, broadly inheriting its tripartite form, since each of the texts represents a different version of the Dido myth: this structure had a significant impact on the musical form of the piece. The tripartition provided a narrative and thematic basis, guiding the flow and evolution of the work, which is articulated as a series of ‘melologues’ in which the voice alternates with the ensemble,\footnote{The premiere of the piece, which took place on 24 July 2024 at the “Nino Rota” Auditorium in Bari, was entrusted to the cellists Roberta Pastore, Donatello Notarnicola, Emanuela Storelli and Fedora Palladino, with sound direction by Daniele Annese, all students of the “N. Piccinni” Conservatory of Bari, whom we particularly thank for their availability and dedication. The composer completed the ensemble as electroacoustic performer.} the latter accompanying the moments of declamation in a more static manner: the two actors are equal partners, whose respective interventions create a constantly evolving dialogue. Each section of the text, inspired respectively by the character of the events as narrated by the composers Niccolò Piccinni, Domenico Sarro and Henry Purcell, is associated with a distinct musical identity, so as to create a dialogue between word and sound that emphasizes the specific emotions and atmospheres of each iteration.

The first melologue, inspired by Piccinni, introduces a shadow of hope and resilience, with music that seeks to balance melancholy with glimmers of light. Here the cellos explore cantabile melodies and euphonic lines that suggest a possibility of rebirth. Although the hope is fragile and barely hinted at, it offers a less definitive closure than the subsequent sections, creating a narrative arc that does not end in pure tragedy but leaves room for an uncertain yet potentially bright future.

The second melologue, based on Sarro, adopts a more dynamic and contrasting musical approach, paralleling the anger and pain expressed in the text. Driving rhythms and pronounced dissonances characterize this part, with the cellos alternating moments of sonic violence with passages of quiet reflection, symbolizing Dido’s inner turmoil as she watches her world collapse. This section embodies the tension and conflict, both inner and outer, that define the character in question.

The third melologue, inspired by Purcell, is characterized by a vocal tone that has by now weakened, reflecting the protagonist’s despair and resignation. The music here uses rarefied harmonies to evoke a sense of inevitability and loss, mirroring the elegiac tone of the text. The scoring is intimate, with the four cellos weaving a sonic fabric that tends to fade, like a distant memory.

Since there will be an opportunity later to dwell on the instrumental writing of IHM, a number of important details should be addressed here. The inherited tripartition is not simply a formal division but a framework that shaped the entire musical conception of the piece. The voices are not physically present on stage but are projected by another actor on stage – besides the cellos – namely the electroacoustic performer, who is also responsible for the real-time processing of the acoustic instruments. He is entrusted with two main cadenzas, the first on fixed media and the second improvisatory in character, which serve as interludes. The core point has already been stated, but it is worth stressing again: the interaction among the actors on stage, human and artificial, constitutes the very dramaturgy of the piece, an essential component of what was once rightly defined as an “\textit{action collective}”. The compositional work was multifaceted and was accompanied by a continuous process of skill acquisition – the true mission of any educational act – which made it possible to maintain a strong awareness throughout all creative phases, following an exploratory method of continuous research or, as Enzo Mari would put it, of ‘\textit{autoprogettazione}’ (self-design).\footnote{Cf. MARI, E., \textit{Autoprogettazione?}, Corraini Edizioni, Mantua 2002.}

\section{IHM: a focus on the instrumental writing}\label{sec:4}

\subsection{Techniques of mimesis}\label{sec:4.1}

Mimesis is a process of imitation or reflection, used in various artistic contexts to represent or reinterpret reality.\footnote{Cf. \textsc{Gebauer, G., Wulf, C.}, \textit{Mimesis: Culture, Art, Society}, University of California Press, Berkeley 1995.} In music, this concept takes shape in the imitation of melodic, rhythmic or timbral motifs between instruments or within a single instrumental group. Mimesis can serve to create a dialogue between voices, to develop a theme or to generate variations that enrich the structure of the piece.

\subsubsection{On the instrumentation}\label{sec:4.1.1}

In \textit{In Human Memories} the choice of four cellos is fundamental to exploring ‘mimesis’ between instruments. The ensemble is unusual and unconventional, departing from the more traditional formations of chamber music. This choice makes it possible to explore timbral and dynamic territories suited to contemporary music, exploiting the homogeneity and versatility of the cello. The absence of differentiated instrumental registers and timbres pushes the composer to exploit expressive nuances, recreating evocative sound environments in the service of the dramaturgy and of the themes outlined above. The whole thus becomes a creative laboratory, where building relationships between instruments, and between instruments and the other actors, requires diversified solutions.

The cellos, with their warm and versatile voice, allow continuous sonic interaction and integration. ‘Mimesis’ among the four instruments is a central element of the piece, creating a musical fabric in which sounds mirror one another and lines chase and merge with each other. This imitative dialogue makes it possible to develop a complex expressive language in which the voices intertwine and influence one another. The four instruments thus become an omnipresent ‘super-instrument’, endowed with its own evocative capacity and supporting the interventions of the pre-recorded and synthesized voices, with a slight reference to the Japanese \textit{shamisen}, a chordophone used to accompany popular \textit{kabuki} theater performances.\footnote{Cf. \textsc{Kodama, S.}, \textit{The Complete Guide to traditional Japanese}, Kodansha International, Tokyo 2000.} From a certain point of view, however, the theatrical arrangement is reversed, since the voices – the intelligible and immediately decodable element – are virtual and projected through ‘holophony’ techniques, which will be touched upon at the end of the paper, while the instruments are always present on stage, in a duality that suggests a form of reunion or even of opposition: an ethereal part and a tangible one, emergence and latency, Dido and Aeneas.

\subsubsection{Heterophony as ‘simultaneous otherness’}\label{sec:4.1.2}

Among the techniques that can be traced back to instrumental mimesis, heterophony certainly plays a prominent role. It is a compositional or improvisational technique in which a single melodic line is performed simultaneously by two or more instruments or voices, but with rhythmic, ornamental or amplifying variations.\footnote{Cf. \textsc{D’Indy, V., Sérieyx, A.}, \textit{Cours de composition musicale. 2E Liv., 1e PT}. A. Durand et fils, 1912, pp. 435-487.} Unlike counterpoint, in which the voices move interdependently, in heterophony all the parts follow the same melodic ‘class’, while subtly diverging from it. This type of texture creates a unique sonic complexity, in which the original melody is enriched by slight dissonances and nuances resulting from the different parts. This technique of ‘simultaneous otherness’\footnote{Cf. \textsc{Pärtlas, Ž.,} \textit{Theoretical Approaches to Heterophony}, “Res Musica”, VIII, 2016.} is common in many non-Western musical traditions, such as the traditional music of the Far East and the Middle East and some practices of sub-Saharan Africa, where improvisation and melodic ornamentation are key elements. Yet heterophony is also present in Western music, albeit less explicitly, especially in the medieval and Renaissance repertoire. Modern and contemporary composers have taken up this concept, integrating it into orchestral or chamber contexts to create evocative, layered sonic results. Heterophony thus represents an interesting meeting point between unison and polyphony, where unity and diversity coexist in a delicate balance, giving rise to a musical fabric rich in details and nuances that emerge from the superimposition of similar, but never identical, voices.

In the piece under examination, heterophony plays a crucial role in the musical fabric, helping to create an iridescent sonic dimension. Heterophony explores a single melodic idea, interpreted differently by each cello, giving rise to a polyphony of subtle variations, as can be seen in the score excerpt shown in Fig.~\ref{fig:4}: indeed, the choice of several instruments of the same ‘class’ is directly connected to the multiplicity of Didos evoked.
\begin{figure}[htbp]
\centering
\includegraphics[width=10cm, trim=52.8bp 75.6bp 42.7bp 422.9bp, clip]{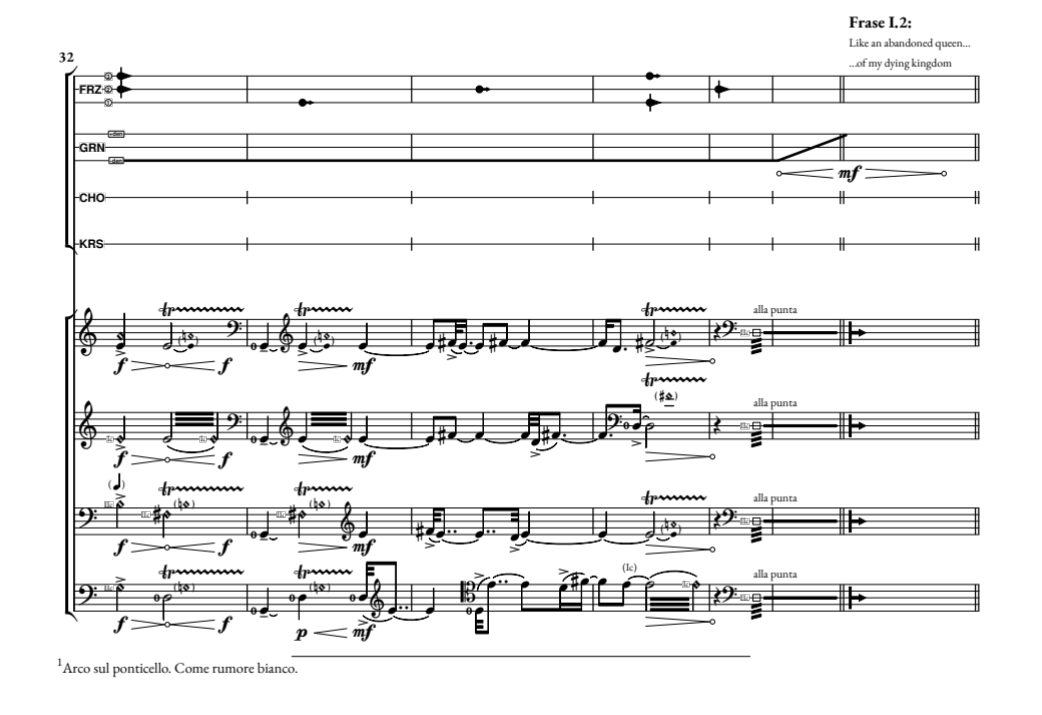}
\caption{Heterophony among the four cellos: the same melodic principle realized in four slightly different ways.}\label{fig:4}
\end{figure}

Other forms of joint participation in the realization of a complex result, although not definable as strictly heterophonic, are present throughout the work, suggesting that the instrumental component takes the form of a sort of ‘super-instrument’ whose combination of interventions creates differentiated listening planes. One example among others can be found at the opening of the third melologue. It takes up sonorities and materials directly from \textit{Dido’s Lament}, Dido’s final aria in Henry Purcell’s opera \textit{Dido and Aeneas}, on which GPT’s textual elaboration is also based. As can be seen in Figure~\ref{fig:5}, each cello carries out a \textit{freeze} process, ‘freezing’ one of the pitches of the famous ground bass and effectively placing it in a virtual reverberant space created by the acoustic instruments alone.

This environment is then populated by the opening material of the aria, giving rise to a sort of round that lets it resound several times, delicately emerging from the background listening plane – the \textit{freeze} process continues simultaneously – and sinking back into it, as can be seen in Figure~\ref{fig:6}.
\begin{figure}[htbp]
\centering
\includegraphics[width=10cm, trim=50.1bp 474.2bp 42.0bp 67.0bp, clip]{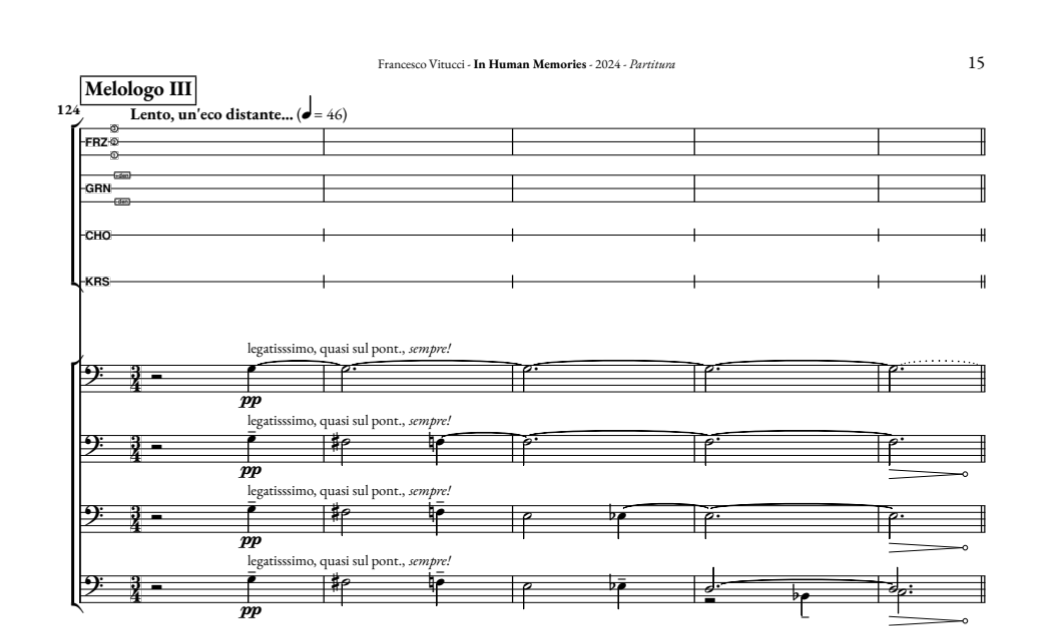}
\caption{Instrumental opening of Melologue III. Each instrument ‘freezes’ one pitch of the ground bass, sustaining it over time.}\label{fig:5}

\bigskip
\includegraphics[width=10cm, trim=50.8bp 94.1bp 41.4bp 450.9bp, clip]{img/score/fig5-6-score.pdf}
\caption{Continuation of the ‘freeze’ process on the musical material.}\label{fig:6}
\end{figure}

It can thus be inferred that the construction of a musical texture plays a central role in the piece, giving complexity and depth to the composition. This implies that several instruments participate simultaneously in a rich, polyphonic sonic fabric. The timbral, rhythmic and dynamic differences between the instruments accentuate the expressive nuances of the main line, enriching the listener’s perception and acting as a structural mechanism that allows the relationships between the individual instances of the ‘super-instrument’ to be explored.

The chosen compositional approach makes it possible to maintain internal coherence despite performance differences between the instruments, an essential compositional device in the search for unity in multiplicity.

\subsubsection{Live electronics: not just a ‘consequent’}\label{sec:4.1.3}

The piece features another actor that interacts, to varying degrees, in a form of ‘chamber music-making’ in the traditional sense: the electroacoustic performer. He is present on stage and is the \textit{kybernetes}\footnote{Helmsman, ship’s pilot. Cf. \url{https://www.grecoantico.com/dizionario-greco-antico.php?lemma=KYBERNHTHS100}} of the entire sonic dramaturgical action, since he is responsible both for projecting the vocal interventions, timed according to the indications in the score, and for processing the sound of the cellos in real time, controlling an algorithmic part of his own of considerable complexity – in other words, a genuine instrument in dialogue with the other actors already introduced.

Without going into the technical details of the implementation, which lie beyond the scope of this paper, one significant aspect should be emphasized. The entire work was conceived so that the presence of the electroacoustic performer could bring certain things to light. First, as already mentioned, this presence is functional and necessary for triggering synchronization processes among performers, a mechanism widely investigated in psychological and neuroscientific studies that mediates group musical practice:\footnote{Cf. \textsc{Cheng, S., Wang, J., Luo, R., \& Hao, N}., \textit{Brain to Brain Musical Interaction: A Systematic Review of Neural Synchrony in Musical Activities}, “Neuroscience \& Biobehavioral Reviews”, CLXIV, 2024.} the live-electronics performer, almost always placed in distant areas for optimal listening, is here present on stage, ‘breathing’ with the acoustic instruments and observing and following their gestures, while sound direction is left to a dedicated figure with whom the performance strategies for shaping the listening experience were discussed in advance. One of the points made in various ways throughout this paper – and which we wish to stress here in conclusion – is that “ensemble musical performance is a typically collective affair”.\footnote{\textsc{Cazzato, A.,} \textit{La didattica collettiva per quattro violini come complemento all’insegnamento strumentale}, “Ex Chordis”, I, 2023, p. 15 (translation by the authors).}

Second, the role of processor, which has always been connected with the idea of live electronics, implies the passage of an ‘activating’ audio signal through a suitably designed and implemented algorithmic zone. Logically, the absence of such a signal, for example during rests of the acoustic instruments, drags the live electronics down to the same level, whereas at those moments it could find its own asynchronous expressive spaces; instead it remains in an ‘antecedent–consequent’ relationship that, by virtue of the real-time paradigm, makes the interventions synchronous. This notion of dependence drove the entire work, pushing the creative team to find solutions to this problem while making them functional to the dramaturgy under construction. Two interludes entrusted to the electroacoustic performer were therefore conceived, separating the sections in which the live electronics tends to act as a consequent and giving them the function of cadenzas: the first is purely acousmatic (deferred-time paradigm, Fig.~\ref{fig:1}), creatively using the vocal fragments of the entire piece, by now completely unrecognizable and bent into pure sound events; the second cadenza is for live electronics, exploiting a series of algorithmic sound-generation strategies based on ‘formant filters’,\footnote{Cf. \url{https://ccrma.stanford.edu/~jos/pasp/Formant_Synthesis_Models.html}} which generate voice-like events, varied by the electroacoustic performer alone, who, in a highly improvisatory form, regains his independence.

\subsection{Notes on the set-up}\label{sec:4.2}

What kind of event does the audience experience? For a number of reasons, briefly addressed below, the piece was conceived and developed for listening in a concert hall. Some features of the sound diffusion system were customized for the concert hall used (and can be adapted to other listening spaces), confirming the intention to realize a compositional idea with site-specific traits,\footnote{Cf. \textsc{Di Scipio, A.}, \textit{Circuiti del tempo}, cit., p. 411.} into which the listener is immersed to live a musical experience that is certainly emotional and at times cathartic.

\subsubsection{Building a sound scene}\label{sec:4.2.1}

The capture of the instrumental signal and its real-time processing, the distinctive feature of live electronics, have been mentioned several times – not to mention, of course, what is known as ‘transparent amplification’ or sound reinforcement, which allows acoustic instruments to reach a higher sound level, yet in an extremely natural way. This is generally achieved through a configuration of one or more microphones – other types of transducers are also often used for more specific reasons – and a configuration of one or more loudspeakers, the terminal elements of the electroacoustic chain, arranged thoughtfully and in any case in relation to the intended idea of diffusion.

Regarding this last aspect, the specific compositional need to reify virtual actors (the voices, \textit{in primis}) on stage led to a series of choices concerning the diffusion system to be used: a three-dimensional listening system. Historically, the search for experiences of this kind dates back to the first experiments in the mid-twentieth century. Pioneering figures such as Michael Gerzon and Peter Fellgett, as early as the 1970s,\footnote{Cf. \textsc{Gerzon, M. A.}, \textit{Periphony: With-height sound reproduction}, “Journal of the Audio Engineering Society”, 21(1), 1973.} laid the foundations for the technique now known as Ambisonics, with the aim of creating a system capable of accurately reproducing three-dimensional sound fields. This system is based on the principle of capturing and reproducing sound fields. By encoding sound sources and their spatial characteristics by means of spherical harmonics, Ambisonics allows precise spatial localization and the manipulation of audio objects within a three-dimensional space.

An immersive technique requires specific listening environments, purpose-designed to allow the physical immersion of the listener in diffusion systems that surround them. It was therefore decided to adopt a listening system that, while guaranteeing the three-dimensionality of Ambisonics, could be arranged on a sound front positioned directly on the stage itself. Five S.T.ONE loudspeakers were used – a type of tetrahedral loudspeaker conceived and developed by Giuseppe Silvi – together with a purpose-written ‘holophony’ software system: this is an advanced frontier in three-dimensional audio reproduction, aimed at recreating the listener’s natural auditory experience by means of a loudspeaker array based on wave field synthesis.\footnote{Cf. BERKHOUT, A. J., DE VRIES, D., VOGEL, P., \textit{Acoustic control by wave field synthesis}, “The Journal of the Acoustical Society of America”, 93, 5, 1993.} Photographic documentation of the set-up is available in the appendix.

\subsubsection{Instrumental and vocal recording techniques}\label{sec:4.2.2}

\begin{figure}[htbp]
\centering
\setlength{\tabcolsep}{0pt}
\begin{tabular}{|>{\centering\arraybackslash}m{5cm}|>{\centering\arraybackslash}m{5cm}|}
\hline
\rule{0pt}{5.1cm}\includegraphics[width=3.08cm]{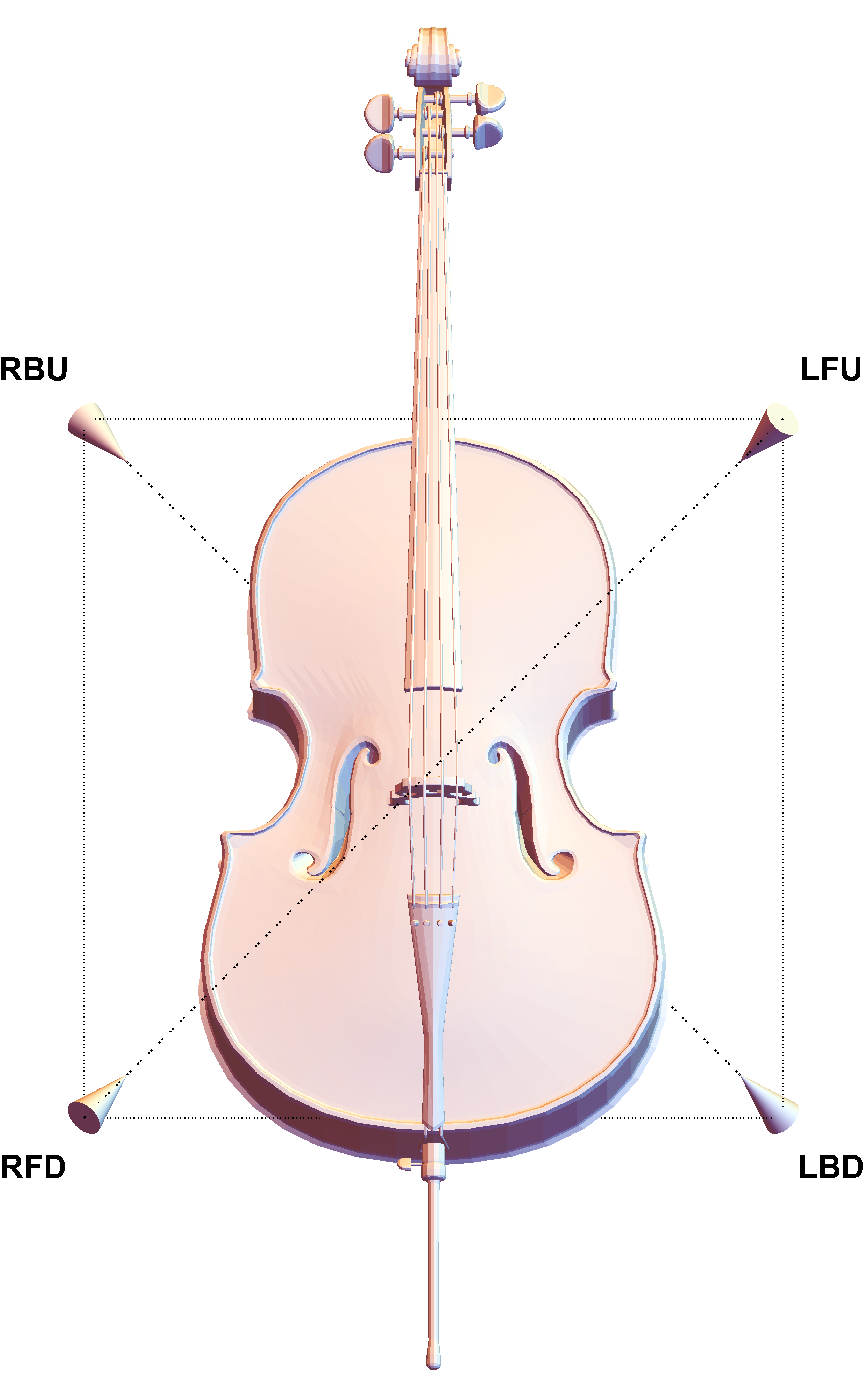} &
\includegraphics[width=3.08cm]{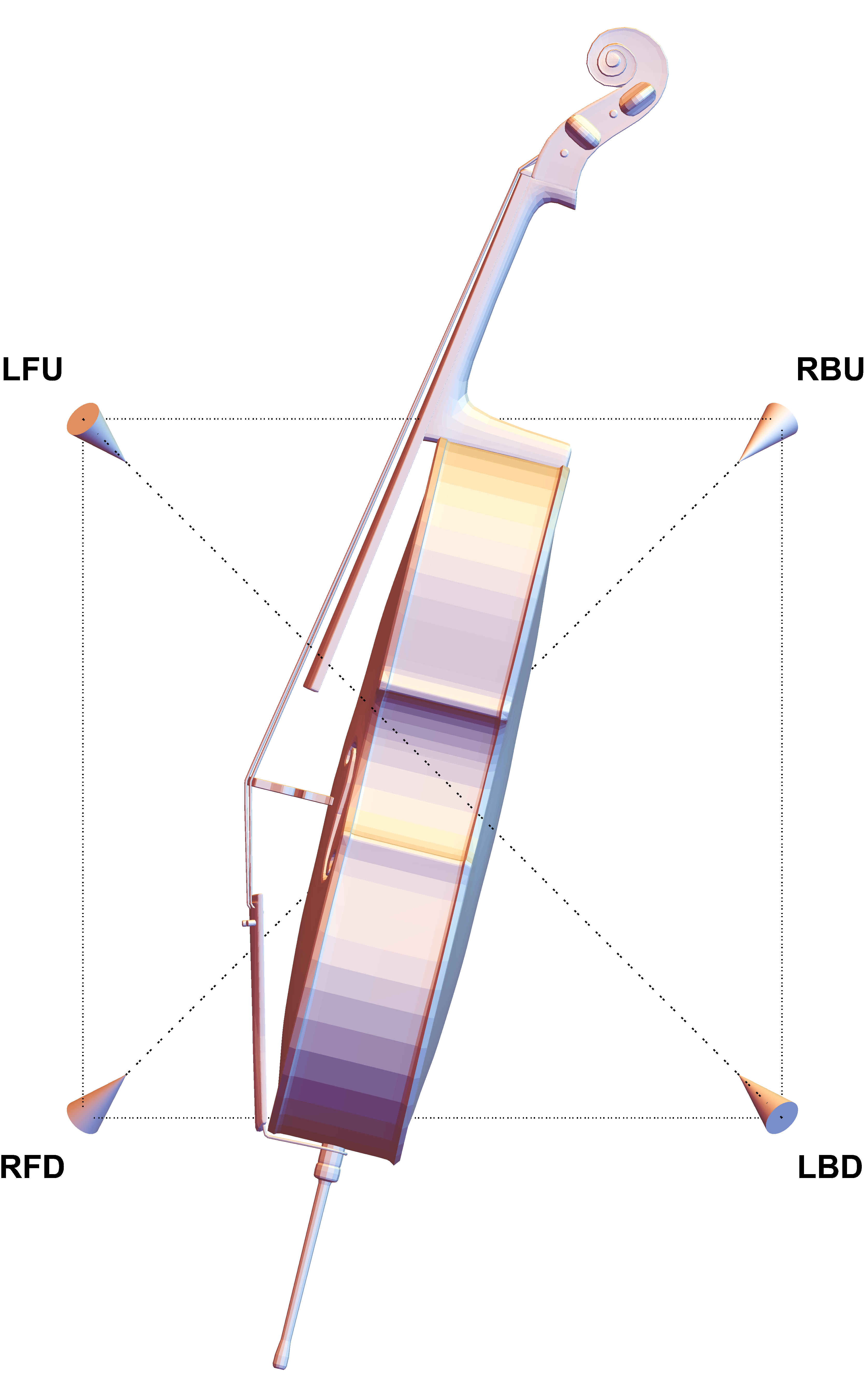} \\[2pt]
\hline
\rule{0pt}{5.1cm}\includegraphics[width=4.26cm]{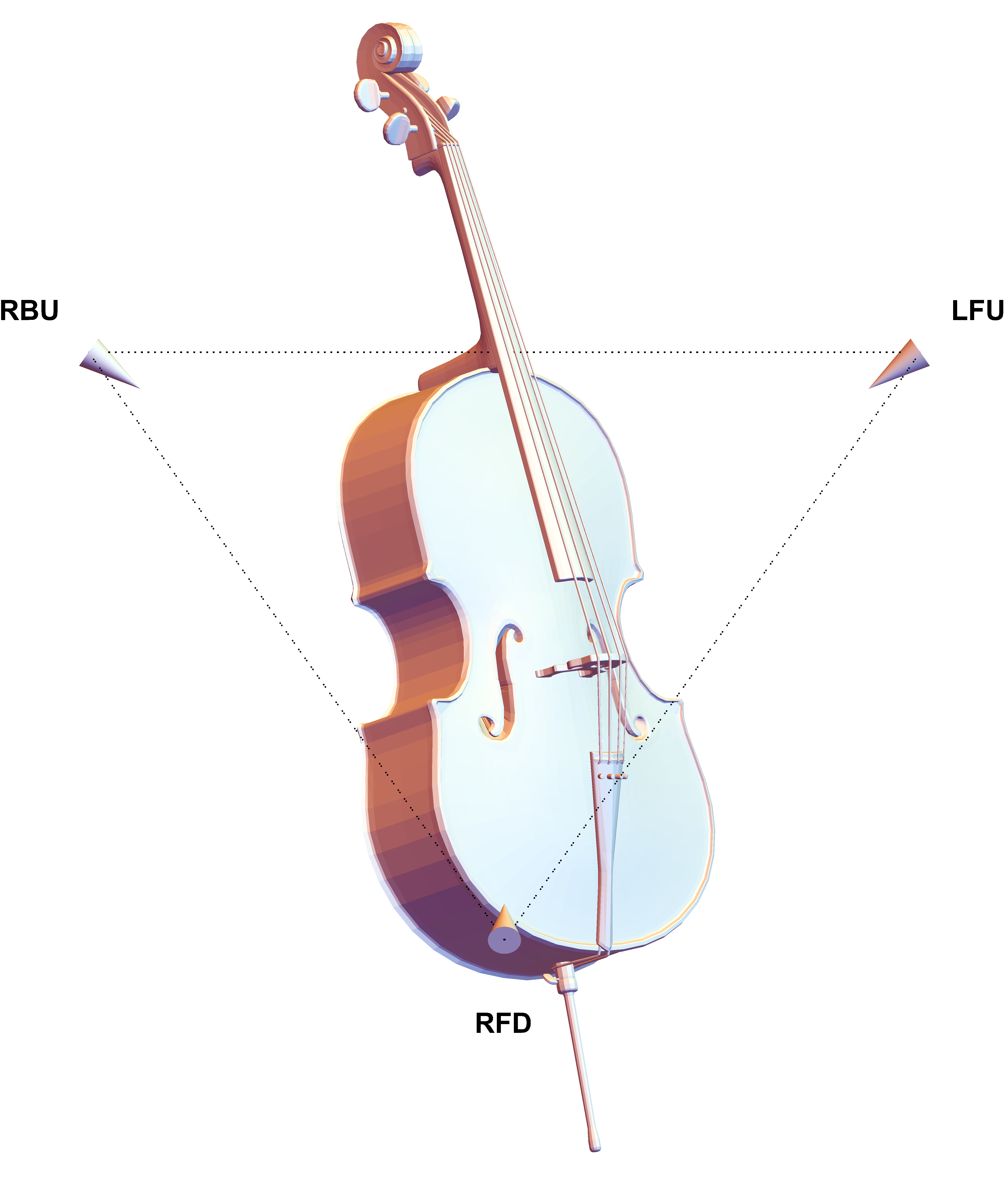} &
\includegraphics[width=4.33cm]{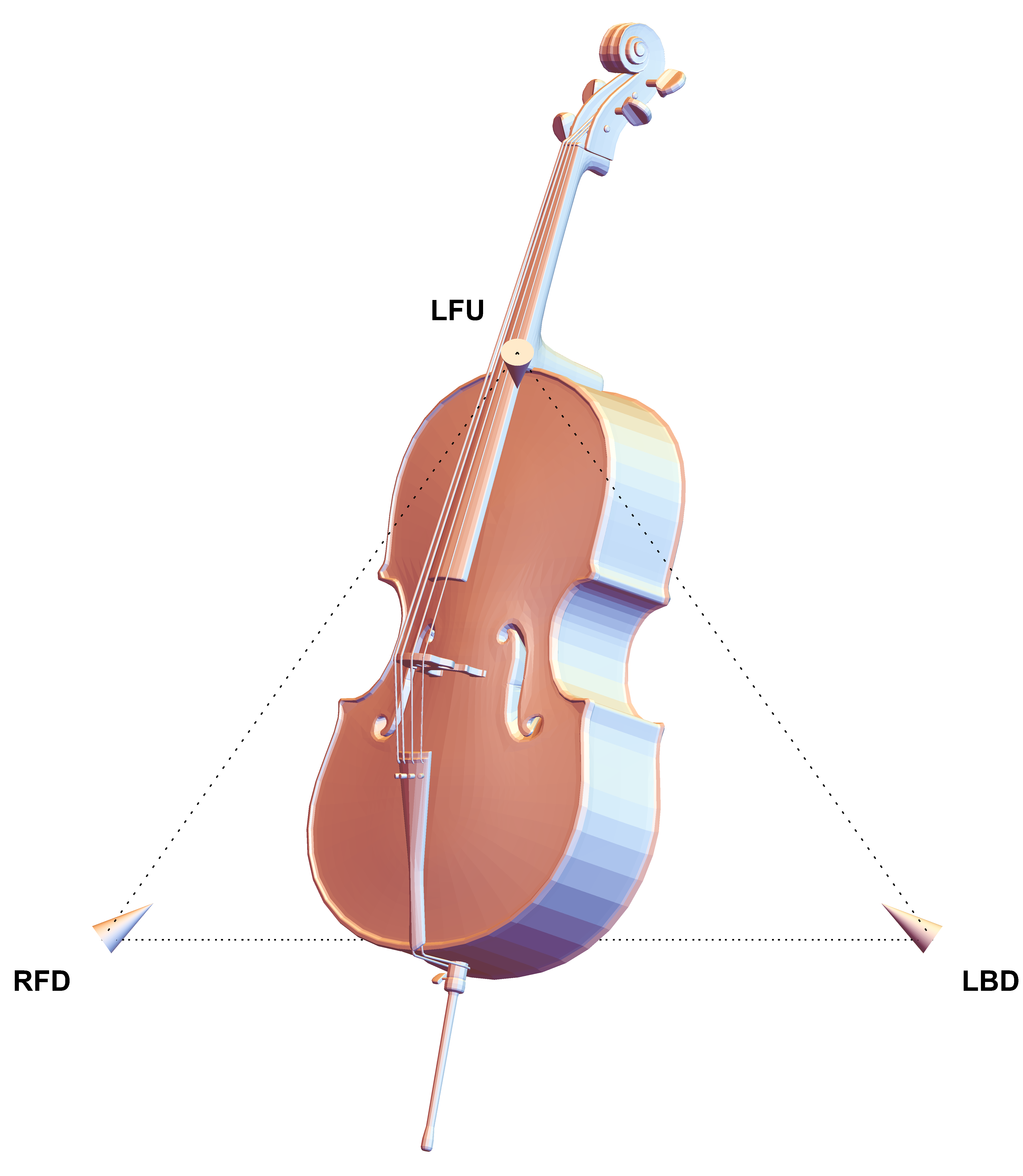} \\[2pt]
\hline
\end{tabular}
\caption{Perspective views of a Tetrarec recording set-up applied to a cello.}\label{fig:7}
\end{figure}

The use of non-standard loudspeakers naturally informed and conditioned all the processes that involved, at any level, the capture of instrumental and vocal sound. The S.T.ONE loudspeaker tetrahedron is thus fed by four signals picked up by a microphone configuration of the same shape. Figure~\ref{fig:7} shows, from different angles, an explanatory three-dimensional reconstruction of this technique, named Tetrarec by the designer of S.T.ONE himself, applied to a cello. The four microphones, whose labels refer to their position relative to the instrument,\footnote{\textbf{L}eft \textbf{F}ront \textbf{U}p, \textbf{R}ight \textbf{F}ront \textbf{D}own, \textbf{R}ight \textbf{B}ack \textbf{U}p, \textbf{L}eft \textbf{B}ack \textbf{D}own.} work together to capture a three-dimensional sound field. On the one hand, the performer can thus interact very freely with the instrument, without worrying about being in the ‘focus’ of a traditional capture system; on the other, this technique natively returns a three-dimensional image, fully compatible with Ambisonics and with the custom systems designed and built for the piece. This recording technique thus enhances each performer’s listening point, transferring it directly to the loudspeakers: environment, performer and instrument – constitutive aspects of the timbral process – work together functionally with the apparatus built.

\section{Conclusions}\label{sec:5}

The experiences reported in this research, both historical and more recent, speak of a strong osmotic relationship between acoustic-instrumental reality and the potential of computer technologies. The survey has shown how expressive needs turned attention towards technological tools to be developed through experimentation. The overwhelming appearance of Artificial Intelligence is leading everyone to measure themselves, at every level, against it and its potential, which is still largely unknown today. In other words, it poses questions that call for an answer. In the second part of the paper, the two case studies thus offered two alternative visions of the relationship with AI: in \textit{Convergence}, the human being is forced to become a coded parameter in a system governed by AI; the experience of \textit{In Human Memories}, by contrast, rediscovers the idea of technology as a tool with which one can now enter directly into dialogue, while it remains at the service of the composer, who can use it consciously by developing the appropriate competences.

\clearpage
\bibheading{Bibliography}


\begin{references}

\textsc{Bernardini N.}, \textit{Live electronics}, “Nuova Atlantide. Venice: la Biennale di Venezia”, Venice 1986.

\textsc{Berkhout, A. J., de Vries, D., Vogel, P.}, \textit{Acoustic control by wave field synthesis}, “The Journal of the Acoustical Society of America”, 93, 5, 1993.

\textsc{Brown, T. B., Mann, B., Ryder, N., Subbiah, M., Kaplan, J. D., Dhariwal, P., Neelakantan, A., Shyam, P., Sastry, G., Askell, A., ET AL.}, \textit{Language models are few-shot learners}, “Advances in Neural Information Processing Systems”, XXXIII, 2020, pp. 1877-1901.

\textsc{Cazzato, A.}, \textit{La didattica collettiva per quattro violini come complemento all’insegnamento strumentale}, “Ex Chordis”, I, 2023, pp. 14-27, \url{https://doi.org/10.54103/3034-8781/22472}.

\textsc{Cheng, S., Wang, J., Luo, R., \& Hao, N}., \textit{Brain to Brain Musical Interaction: A Systematic Review of Neural Synchrony in Musical Activities}, “Neuroscience \& Biobehavioral Reviews”, CLXIV, 2024.

\textsc{Clark, A.}, \textit{Surfing Uncertainty: Prediction, Action, and the Embodied Mind}, Oxford Academic, online edn., New York, 2016,

\url{https://doi.org/10.1093/acprof:oso/9780190217013.001.0001}.

\textsc{D’Indy, V., Sérieyx, A.}, \textit{Cours de composition musicale. 2E Liv., 1e PT}. A. Durand et fils, 1912.

\textsc{Di Scipio A.}, \textit{Circuiti del tempo: Un percorso storico-critico nella creatività musicale elettroacustica e informatica}, Libreria Musicale Italiana, Lucca 2021.

\textsc{Floridi L.}, \textit{La quarta rivoluzione: come l'infosfera sta trasformando il mondo}, Raffaello Cortina Editore, Milan 2017.

\textsc{Gebauer, G., Wulf, C.}, \textit{Mimesis: Culture, Art, Society}, University of California Press, Berkeley 1995.

\textsc{Gerzon, M. A.}, \textit{Periphony: With-height sound reproduction}, “Journal of the Audio Engineering Society”, 21(1), 1973.

\textsc{Goodfellow I., Bengio Y., Courville A.}, \textit{Deep Learning}, MIT Press, Cambridge 2016.

\textsc{Hiller L., Isaacson L.}, \textit{Experimental Music: Composition with an Electronic Computer}, McGraw-Hill, New York 1959.

\textsc{Khan M. E., Khan F.}, \textit{A comparative study of white box, black box and grey box testing techniques}, “International Journal of Advanced Computer Science and Applications”, III, 6, 2012.

\textsc{Kodama, S.}, \textit{The Complete Guide to traditional Japanese}, Kodansha International, Tokyo 2000.

\textsc{Mari, E.}, \textit{Autoprogettazione?}, Corraini Edizioni, Mantua 2002.

\textsc{Menger P. M.}, \textit{Les Laboratoires de la création musicale}, La Documentation française, Paris 1989.

\textsc{Oord, A. v. d., Dieleman, S., Zen, H., Simonyan, K., Vinyals, O., Graves, A., Kalchbrenner, N., Senior, A., \& Kavukcuoglu, K.}, \textit{WaveNet: A generative model for raw audio}, “Proceedings of the 9th ISCA Speech Synthesis Workshop”, CXXV, 2016.

\textsc{Pärtlas, Ž.,} \textit{Theoretical Approaches to Heterophony}, “Res Musica”, VIII, 2016.

\textsc{Piaget, J.}, \textit{The Construction of Reality in the Child}, Basic Books, New York 1954.

\textsc{Rice A. R.}, \textit{Notes for Clarinetists: A Guide to the Repertoire}, Oxford University Press, Oxford 2016.

\textsc{Vitucci F., Silvi G., Annese D. G., Scagliola F., Di Furia A.}, \textit{Opening mind by opening architecture: analysis strategies}, “7\textsuperscript{th} International Csound Conference Proceedings”, 2024.

\textsc{Zattra L.}, \textit{Les origines du nom de RIM (Réalisateur en informatique musicale)}, “JIM – Journées d’Informatique Musicale”, Saint-Denis, Paris 2013.
\end{references}
\end{document}